\documentclass[10pt,twocolumn,letterpaper]{article}

\usepackage{cvpr}
\usepackage{times}
\usepackage{epsfig}
\usepackage{graphicx}
\usepackage{amsmath}
\usepackage{amssymb}

\usepackage[breaklinks=true,bookmarks=false,colorlinks,linkcolor=red]{hyperref}

\cvprfinalcopy 

\def\cvprPaperID{****} 

\ifcvprfinal\fi
\begin{document}

\title{Deep Video Matting via Spatio-Temporal Alignment and Aggregation}

\author{Yanan Sun\\
HKUST\\
{\tt\small now.syn@gmail.com}
\and
Guanzhi Wang\thanks{The second and third authors contribute equally to this work.}\\
Stanford University\\
{\tt\small guanzhi@cs.stanford.edu}
\and
Qiao Gu$^*$\\
Carnegie Mellon University\\
{\tt\small qiaog@andrew.cmu.edu}

\and
Chi-Keung Tang\\
HKUST\\
{\tt\small cktang@cs.ust.hk}

\and
Yu-Wing Tai\\
Kuaishou Technology\\
{\tt\small yuwing@gmail.com}

}

\maketitle

\begin{abstract}
\vspace{-5pt}
Despite the significant progress made by deep learning in natural image matting, there has been so far no representative work on deep learning for video matting due to the inherent technical challenges in reasoning temporal domain and lack of large-scale video matting datasets. In this paper, we propose a deep learning-based video matting framework which employs a novel and effective spatio-temporal feature aggregation module (ST-FAM). As optical flow estimation can be very unreliable within matting regions, ST-FAM is designed to effectively align and aggregate information across different spatial scales and temporal frames within the network decoder. To eliminate frame-by-frame trimap annotations, a lightweight interactive trimap propagation network is also introduced. The other contribution consists of a large-scale video matting dataset with groundtruth alpha mattes for quantitative evaluation and real-world high-resolution videos with trimaps for qualitative evaluation. Quantitative and qualitative experimental results show that our framework significantly outperforms conventional video matting and deep image matting methods applied to video in presence of multi-frame temporal information. Our dataset is available at {\small \url{https://github.com/nowsyn/DVM}}.
\end{abstract}
{\let\thefootnote\relax\footnote{{This work was done when Yanan Sun was a student intern at Kuaishou Technology. This work was supported by Kuaishou Technology and the Research Grant Council of the Hong Kong SAR under grant no. 16201420.}}}

\vspace{-15pt}
\section{Introduction}
\vspace{-2pt}
Video matting, or extracting from a given video high-quality alpha matte of a moving foreground object, has a wide range of applications in special effect and TV/movie production. Formally, given the color of a video frame $I$,  foreground color $F$,  background color $B$ and  alpha matte $\alpha \in [0,1]$, the video compositing equation Eq.~\ref{eq:alphamatting} is
\begin{equation}\label{eq:alphamatting}
    I = \alpha F + (1-\alpha) B.
\end{equation}
Compared to image matting, video matting poses two further challenges. First, video matting needs to preserve spatial and temporal coherence in the predicted alpha matte.
A straightforward solution  applying image matting on  individual frames may inevitably cause severe flickering artifacts for moving fine details. Using optical flow to regularize output may help to alleviate these artifacts, but even with the most state-of-the-art optical flow estimation methods~\cite{ilg2017flownet, sun2018pwc, zhao2020maskflownet}, optical flow estimation within complex matting regions is still very unreliable. This is because  matting regions simultaneously contain both the foreground and background and there is so far no good optical flow estimation that can handle large area of semi-transparency. 
\begin{figure}[t]
\centering 
\includegraphics[width=1.0\linewidth]{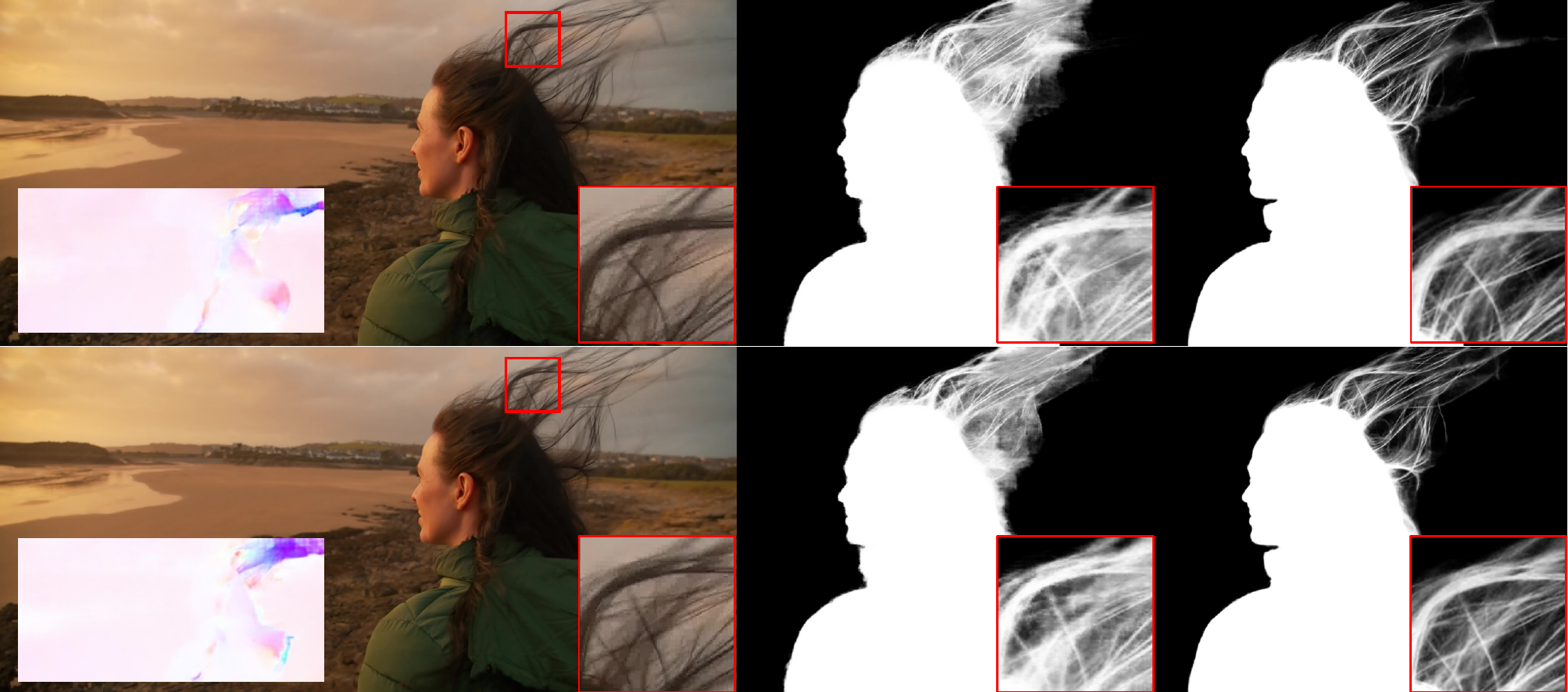} 
\caption{A challenging video matting example where top and bottom are consecutive frames. Left: input frames with insets zooming in complex hairs and showing the estimated optical flow from PWC-Net~\cite{sun2018pwc}. Note that the estimated optical flow is unreliable within the hair regions. Middle: deep image matting~\cite{Xu2017DeepIM} results. Right: our results.  }
\label{fig:teaser}
\vspace{-0.15in}
\vspace{-2pt}
\end{figure}

Traditional methods tackled the video matting problem by finding local or non-local affinity among pixel colors and computing the motion of the foreground~\cite{Choi-2013-videomattingmultiframe,Zou2019sparselowrank} but their results are still far from satisfactory especially when dealing with complex cases, such as rapidly moving objects or complex backgrounds. Figure~\ref{fig:teaser} shows an example with challenging motions. The other challenge for video matting is the necessary input of a dense trimap for each frame, making it difficult to generate high quality large-scale video matting benchmarks.

In this paper, we propose an  encoder-decoder network consisting of a novel spatio-temporal feature aggregation module (ST-FAM) for extracting feature pyramids at different levels, which utilizes spatial and temporal information across multiple frames. {\em Without} optical flow estimation, our network can effectively address the video matting problem and produce spatially and temporally coherent alpha mattes, and can generate good predictions of hard cases using the fused temporal information. To provide reliable frame-by-frame trimaps with minimum user inputs, a novel correlation layer is introduced to propagate trimaps across different frames. With our trimap propagation method, a user can edit and propagate trimaps at an interactive frame rate. 

To support our and future video matting research, we have also contributed a high-quality video matting dataset with groundtruth alpha mattes. Furthermore, to verify the generalization capability of our method to real videos, we provide 10 high-resolution real-world videos with dense and frame-by-frame human annotated trimaps for evaluation. We evaluate our method on our composited test set as well as real-world high-resolution videos. Experimental results demonstrate that our deep video matting method significantly outperforms image-based deep matting methods and conventional video matting approaches, capable of handling complex scenarios such as rapidly moving objects with fuzzy boundaries or complex backgrounds.

\vspace{-2pt}
\section{Related Work}

\subsection{Image Matting}
Traditional methods on natural image matting mainly use color and other relevant low-level image features for estimating alpha matte via sampling, propagation or a combination of both. Sampling-based methods~\cite{Chuang-2001-CVPR-bayesianmatting,Feng-2016-ECCV-clustersampling,GastalOliveira-2010-CGF-SharedMatting,He-2011-cvpr-globalsampling,Ruzon-2000-cvpr-alphaestimation} 
first sample pixels from the foreground and background in a given image to construct pertinent color models which are used to estimate alpha values in the transition region. In propagation-based methods~\cite{Aksoy-2017-cvpr-ifm,Aksoy-2018-tg-sss,bai-iccv-geodesicframework,Chen-2012-PAMI-knnmatting,grady2005random,levin2008closed,levin2008spectral}, Eq.~\ref{eq:alphamatting} is reformulated so that alpha values are allowed to propagate from known foreground and background into the unknown transition region. Please refer to \cite{wang2008mattingsurvey} for a comprehensive review on traditional matting methods. 

For deep-learning based image matting,
Cho~\etal~\cite{Cho-2016-mattingusingdeepcnn} proposed to apply deep neural networks to  combine the complementary advantages of the results respectively produced by closed-form matting~\cite{levin2008closed} and KNN matting~\cite{Chen-2012-PAMI-knnmatting}.
Xu~\etal~\cite{Xu2017DeepIM} proposed a two-stage encoder-decoder network followed by a refinement network to address the image matting problem.  
Lutz~\etal~\cite{Lutz2018AlphaGANGA} proposed a generative adversarial network where dilated convolutions are integrated into the encoder-decoder network for improving matting performance. 
Wang~\etal~\cite{wang-ijcai2018-deeppropagation} introduced deep neural networks to learn an alpha matte propagation principle. 
Recently, a number of works have been proposed focusing on relaxing trimap input.
Shen~\etal~\cite{Shen2016DeepAP} used an average shape mask for portraits to guide the network to infer the foreground and background regions automatically. 
Following the similar idea, Zhu \etal~\cite{zhu2017fastdeepmattingportrait} designed a smaller portrait matting network and a fast filter that can be run in real time. 
Chen \etal~\cite{Chen2018SemanticHM} further eased the need for trimap on human, where a segmentation network is first used to predict foreground, background and transition regions from the input image. These regions are then fed together into another network to predict alpha matte. 
Zhang \etal~\cite{zhang2019latefusioncnn} extended this idea to general objects. They used a CNN with two decoders for foreground and background classification, and the classification results are then fed into a fusion network to obtain the alpha matte. 
Liu \etal~\cite{liu2020boosting} leveraged coarse annotated data with fine annotated data for boosting human matting without trimaps. They applied a mask prediction network taking hybrid data for generating human mask, a quality unification network for aligning the mask, and finally a matting refinement network for predicting the alpha matte. 
Qiao \etal~\cite{qiao2020attention} further improved the performance of general object matting without trimap via attention mechanism. They proposed an end-to-end hierarchical attention network exploiting spatial and channel-wise attention to utilize appearance cues in a novel fashion.
Sengupta \etal~\cite{sengupta2020background} introduced a different input setting. Their framework takes two photos with and without the foreground object as input to  reduce trimaps labeling labor and provide external clues for model at a low cost.

\begin{figure*}[ht]
\centering 
\includegraphics[width=1.0\linewidth]{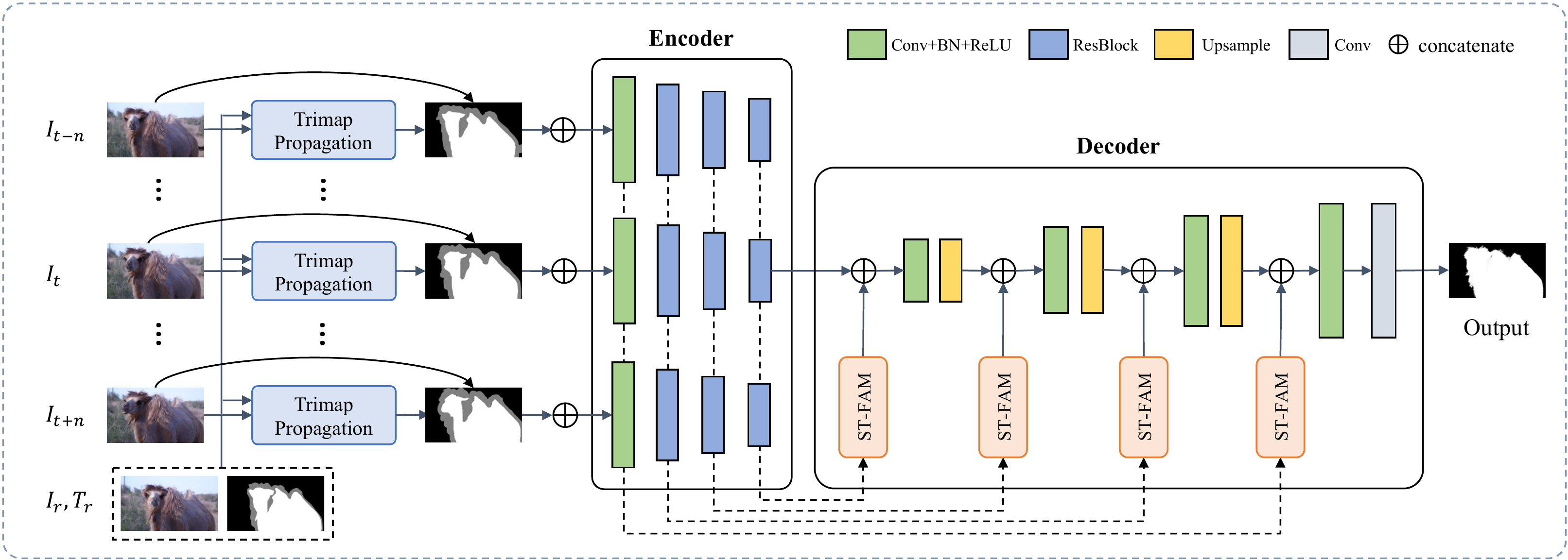} 
\caption{Our video matting framework. The lightweight trimap propagation network generates trimap of target frame $I_t$ according to the trimap of reference frame $I_r$. The spatio-temporal feature aggregation module (ST-FAM) analyzes and fuses information at different levels from neighboring frames $\{I_{t-n},\cdots, I_{t+n}\}$ and progressively outputs aggregated features to be fed into corresponding layers of the decoder. The structures of trimap propagation module and ST-FAM are shown in Figure~\ref{fig:submodule}.} 
\label{fig:framework}\vspace{-0.15in}
\end{figure*}

\vspace{-3pt}
\subsection{Video Matting}
Temporal coherency consideration is important for generating high-quality video matte.
Chuang \etal~\cite{Chuang2002videomattingcomplex} used forward and backward optical flow to interpolate framewise trimaps and applied Bayesian matting to produce high-quality mattes of moving objects. 
Lee \etal~\cite{lee2010temporallycoherentvideomatting} extended robust matting~\cite{Wang2007OptimizedCS} by regarding time as a third spatial dimension and developed an anisotropic kernel using optical flow. 
Bai \etal~\cite{bai2011towardstemporallycoherent} used a temporal matte filter to improve temporal coherence while preserving matte structures on individual frames. 
Choi \etal~\cite{Choi-2013-videomattingmultiframe} used multi-frame non-local matting Laplacian in spatial temporal domain. 
Li \etal~\cite{Li-2013-knnlaplacianvideomatting} implemented motion-aware KNN Laplacian to improve clustering of moving foreground pixels.
Zou \etal~\cite{Zou2019sparselowrank} introduced a sparse and low-rank representation to construct non-local structure which yields better video matting results in terms of spatial and temporal consistency. 
A number of works tackled the video matting problem using custom hardware systems in video capture. 
Neel \etal~\cite{Neel2006cameraarrays} refocused images taken by a camera array to automatically generate alpha mattes for all video frames. 
McGuire \etal~\cite{Mcguire2007defocusvideomatting} computed alpha mattes from synchronized video streams taken by multiple cameras from the same point of view but with varying focus. 

While there exist many traditional methods tackling video matting, their results are not as good as the recent deep learning-based methods, because the deep structural and semantic features are superior to low-level color features in traditional methods. In tackling spatial temporal coherency, later methods~\cite{Choi-2013-videomattingmultiframe,Li-2013-knnlaplacianvideomatting,Zou2019sparselowrank} that utilized non-local matting Laplacian to encode  coherency have demonstrated better performance than the earlier post-processing based methods~\cite{Chuang2002videomattingcomplex,lee2010temporallycoherentvideomatting,bai2011towardstemporallycoherent}. This is because the non-local matting Laplacian not only models the pertinent motions but also  non-local similarities across different patches in different frames. We therefore believe that a deep network that can model the coherency inside its architecture to aggregate different scale of spatio-temporal features should outperform methods using straightforward post-processing.

\section{Datasets}

\subsection{Composited Dataset}
While there exist high-quality and large-scale datasets for image matting~\cite{Rhemann-2009-perceptuallybenchmarkimage,Xu2017DeepIM}, only a few video matting datasets with ground truth alpha mattes are available which are not suitable for training deep neural networks due to their limited sizes~\cite{erofeev2015perceptually}. Thus, we create a new video matting dataset, which is composed of real foreground videos, their groundtruth alpha mattes, and background videos of a great variety of natural and real-life scenes. 

Our foreground objects are made of both images and videos. For foreground video objects, we collect available green screen video clips from the Internet, from which we extract foreground color and alpha matte using a chroma keying software provided by Foundry Keylight~\cite{foundry-keylight}. Since the background of these videos is clean and simple, it is easy to estimate the accurate foreground and alpha matte. Additionally, we also include high-quality images with groundtruth alpha mattes from the Adobe Deep Matting dataset~\cite{Xu2017DeepIM} as foreground images. We discard similar images of the same object and keep 325 images out of 431 training samples and all 50 testing images.

\begin{figure*}[t]
\centering 
\includegraphics[width=1.0\linewidth]{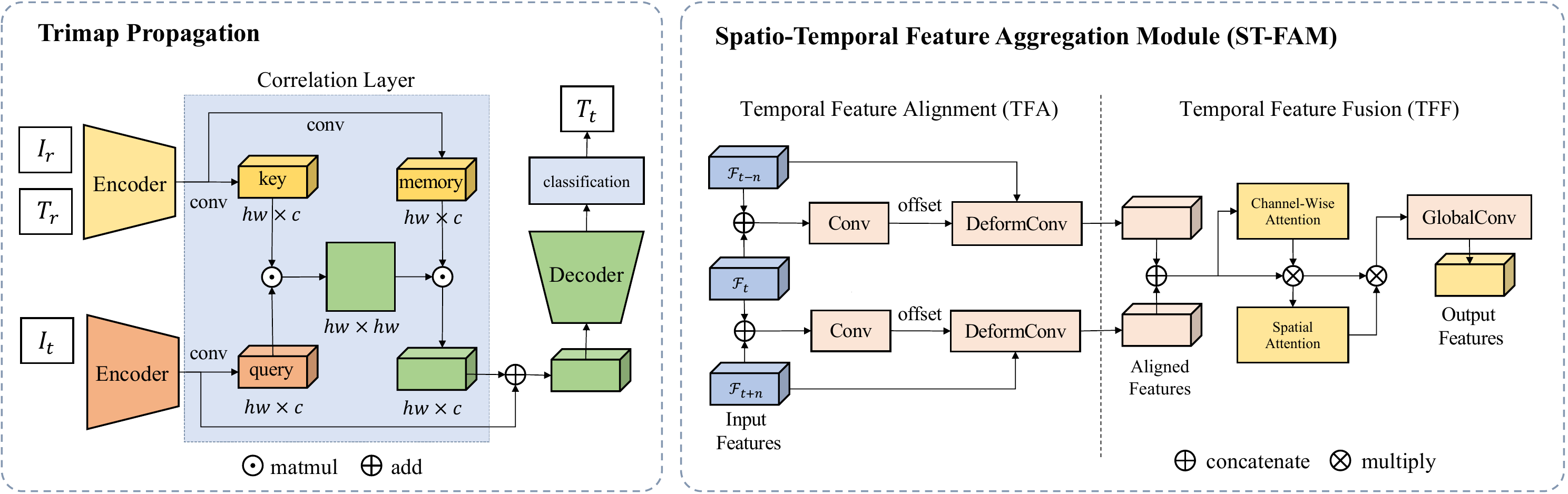} 
\caption{Detailed structure of trimap propagation network and spatio-temporal feature aggregation module (ST-FAM) introduced in Figure~\ref{fig:framework}. In the trimap propagation network, a correlation layer is used to find the mapping of pixels between $I_r$ (reference) and $I_t$ (target); $T_{r|t}$ denotes trimap. ST-FAM is composed of temporal feature alignment (TFA) and temporal feature fusion (TFF), which are respectively responsible for aligning and aggregating features of different frames.}\vspace{-0.15in}
\label{fig:submodule}
\end{figure*}

The background set consists of various real-life videos. We collect over 6500 free video clips of natural scenarios, city views and indoor environment from the Internet. Most of these background videos are HD videos, and a few of them are 4K videos. We composite the foreground video and images onto the background videos using Eq.~\ref{eq:alphamatting}. During the composition, random and continuous translation, rotation and zooming are applied onto the foreground objects to simulate real videos containing a moving foreground. 

Specifically, we composite each foreground object from 325 images and 75 videos with 16 randomly selected background videos, which generates 6400 videos as the train set. For the test set, we similarly combine each object from 50 images and 12 videos with 4 background videos, thus generating 248 test samples. The train and test sets are disjoint. To reduce memory and time cost for training and testing, each composited video contains at most 150 frames with the long side no more than 1920p. Compared to other video datasets, our dataset covers more varieties of video matting and provides rapidly-moving objects, which poses greater challenges for video matting evaluation. 

\subsection{Real-World High-Resolution Videos}
In addition to the composited dataset, we collect 10 real-world videos at 4K resolution to evaluate the generalization ability of our video matting method. These videos are carefully selected from the Internet, consisting of various objects with large motions and complex scenes of real-life including humans, animals, plants, etc. 

\section{Method}
Figure~\ref{fig:framework} shows the overall framework, which consists of a lightweight trimap propagation module and a multi-frame encoder-decoder network. The trimap propagation module predicts the trimap for a target frame given a reference frame with trimap. The encoder extracts features at multiple levels from the input frames and trimaps. The decoder consists of a number of spatio-temporal feature aggregation modules (ST-FAM) which integrate deep features in neighboring frames to enhance alpha prediction of target frame. Figure~\ref{fig:submodule} shows the detailed structures of the two networks.

We use subscript $t$, $r$ to respectively stand for target and reference frame. Here, reference frame is the frame provided with a user-labeled trimap. 

\subsection{Trimap Propagation}
Our trimap propagation method is based on region similarity measures between the reference and target frames. Without computing any optical flow, the network uses two encoders sharing the same structure to respectively extract semantic features of image-trimap $\{I_r,T_r\}$ pair of the reference frame and image $I_t$ of the target frame. We denote the resulting reference feature from the last layer of its corresponding encoder as $\mathcal{F}_r$ and target feature as $\mathcal{F}_t$.
To enlarge receptive fields and overcome different motion as well as scales between the reference and target frames, we apply a cross-attention based correlation layer to match the reference and target frames.

\begin{figure}[t]
\centering 
\includegraphics[width=1.0\linewidth]{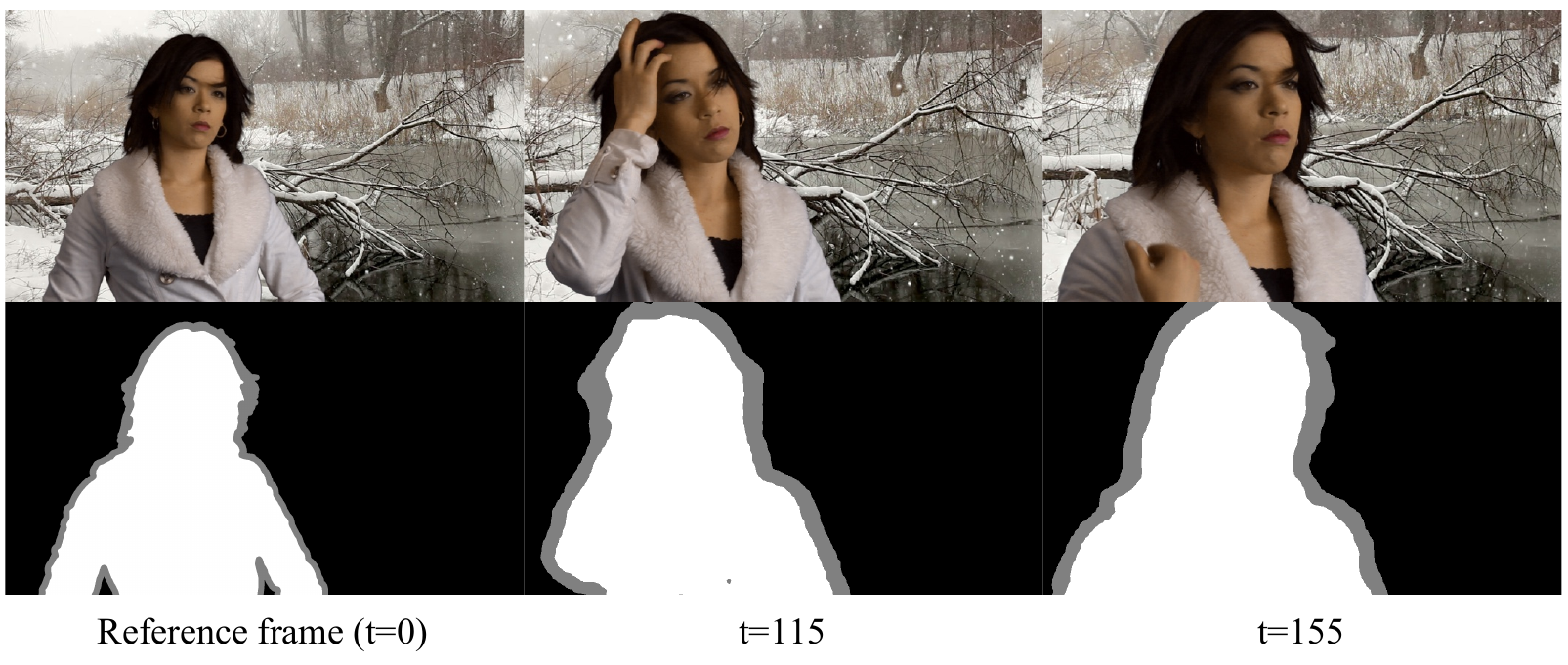} 
\caption{Trimap propagation. The first column is the reference frame  and trimap ($t=0$). The second and third columns are the propagated trimaps at $t=115$ and $t=155$ respectively.}
\label{fig:trimap_setting}
\vspace{-0.15in}
\end{figure}

This correlation layer is composed of key, queries and memories. Key and queries are used to generate correlation scores between the reference and target frames, while memories are applied to enhance correlated features. Given the features $\mathcal{F}_r$ and $\mathcal{F}_t$ of shape $hw\times c$, we adopt Wang~\etal~\cite{Wang_2018_CVPR} to apply a matrix multiplication between queries and keys to get a similarity matrix of shape $hw\times hw$. Intuitively, if a pixel in target frame is highly correlated to a pixel in reference frame, the correlation score between these two pixels will be high. Thus, if a pixel in target frame belongs to the foreground (resp. unknown) region, it should be matched to a corresponding foreground (resp. unknown) pixel in reference frame. These correlation scores are then multiplied with the memory features. The weighted memory features are regarded as residuals and added to $\mathcal{F}_t$. This allows trimap information of reference frames to be propagated to the target frame without affecting the computation of correlation scores. 
Finally, these aggregated features are decoded to classify all pixels into three categories, i.e., foreground, background or unknown, through a classification head.
An example of trimap propagation is provided in Figure~\ref{fig:trimap_setting}.

\subsection{Encoder-Decoder Network}
After generating coarse trimaps for the target frame, our deep video matting framework employs an effective auto encoder-decoder structure to extract features of multiple image-trimap pairs.  
We first apply an encoder network to extract both low-level structural features and high-level semantic features of pixels. Specifically, this encoder network receives multiple frames with corresponding propagated trimaps to extract pyramid features at different levels. We adopt ResNet-50~\cite{he2016deep} as our encoder and collect the features after each residual block. These features obtained at different levels are sent to the decoder for alpha predictions.

In order to predict more accurate alpha matte, a delicate decoder is designed to upscale features from the last layer of the encoder to the same resolution as the input image with several up-convolution layers. We apply sub-pixel convolution layer to upsample features, rather than unpooling operation or deconvolution, for both accuracy and efficiency:  unpooling operation generates sparse indices and sometimes leads to zero gradients, while deconvolution suffers efficiency problem.

Our design is adapted from the U-Net architecture \cite{ronneberger2015u} with skip-connections to preserve both global context features and local detail information. The main difference of our decoder from other U-Net structures lie in the skip-connections, which are enhanced by a novel spatio-temporal feature aggregation module (ST-FAM).

\subsection{Spatio-Temporal Feature Aggregation Module}

The main issues in video matting compared to image matting are how to utilize  temporal information across multiple frames to help distinguish foreground and background color, and how to improve temporal consistency of alpha matte. To achieve these goals, we not only need to consider global context information and local detailed structural information in a single frame, but also need to incorporate motion information of moving pixels by utilizing temporal information from neighboring features to enhance our predictions. To this end, we propose a novel spatio-temporal feature aggregation module (ST-FAM) to exploit information inherent in the features at different scales and timestamps. Figure~\ref{fig:submodule} shows the structure of ST-FAM. Overall, it is composed of temporal feature alignment module (TFA) and temporal feature fusion module (TFF). More implementation details of ST-FAM can be found in supplementary materials.

\vspace{4pt}
\noindent{\textbf{Temporal Feature Alignment Module.}}
The major advantage of videos compared to images in matting is that consecutive frames provide temporal information of foreground objects and background. Motion information of pixels is useful in distinguishing foreground and background colors, which helps the model learn more accurate appearance of foreground objects and handle hard matting cases with complex background. To effectively exploit temporal information, we design a light-weight module that can effectively aggregate temporal features. 

Wang \etal~\cite{wang2019edvr} proposed a pyramid, cascading and deformable module to deal with motions in video restoration task. Inspired by this work, we make our model aware of motion information by aligning the features of neighboring frames with features of target frame. Specifically, for pixel $p$ at time $t$ of frame $I_t$, we try to learn offset $\Delta p$ for $p$ implicitly and translate the offset to obtain aligned features by deformable convolution. Formally, the aligned feature $\mathcal{F}^*$ at time $t+\Delta t\ (\Delta t\in\{0,\pm{1},\dots\})$ of $p$ is defined as
\begin{equation}\label{eq:deform}
    \mathcal{F}_{t+{\Delta t}}^{*}(p) = \sum_{k} w_k \mathcal{F}_{t+{\Delta t}}(p +  {\Delta p_k})
\end{equation}
where $k$ and $w_k$ respectively represent the deformable convolution kernel location and the corresponding weight. $\Delta p_k$ is a {\em learnable} offset from the concatenation of features at time $t$ and $t+\Delta t$. Learning offset and aligning features between $t$ and $t+\Delta t$ enable our model to automatically map identical or similar regions and pixels by their high-dimensional feature expression in temporal context, and consequently to encode temporal information within the aligned features.

\vspace{4pt}
\noindent{\textbf{Temporal Feature Fusion Module.}}
The aligned features obtained by TFA above are passed through the TFF step. As our goal is to obtain fused features for target frame, a simple plausible solution is to compute their average values. However, this will introduce noise or ambiguous information when moving pixels of target frame are lost in neighboring frames. To reduce confusion, our model should pay attention only to relevant information useful for the prediction of target frame. To this end, we introduce an attention mechanism. We perform channel-wise attention as well as spatial attention on the aligned features, by guiding our model to leverage the importance of different channels and the interest regions within a channel. Specifically, we compute a channel attention weight map by applying a global average pooling layer followed with a fully-connected layer on the aligned features, and multiply this map with the aligned features. Then the output feature is multiplied with a learnable spatial attention weight map. Finally, a $1\times 1$ convolution layer and a global convolution layer~\cite{peng2017large} are applied to reduce channels and enlarge receptive fields respectively. 

\vspace{4pt}
After this module, we obtain temporally enhanced features from different blocks which are used in the skip connection with the up-scaled features to fully exploit information from both high-level and low-level features. 
After deriving useful features aggregated in both spatial and temporal dimensions from our decoder, we apply a prediction head, composed of a $3\times3$ convolution and a sigmoid function, to generate the alpha matte for target frame.

\subsection{Loss Functions}
Our network uses multiple losses, including alpha prediction loss and composition loss, which are widely applied in many deep image matting methods. Meanwhile, the gradient loss, KL-divergence loss, and temporal coherence loss are also used.

\vspace{4pt}
\noindent{\textbf{Alpha Loss.}} At timestamp $t$ and pixel $p$, with the alpha matte prediction ${\alpha}_{p,t}$, and the ground truth  $\hat{{\alpha}}_{p,t}$, we define the difference loss of predicted alpha as
\begin{equation}\label{alpha_loss}
    L_a =     
    \begin{cases}
    {||{\alpha}_{p,t}-\hat{\alpha}_{p,t}||_2}, & \text{if}\;{\alpha}_{p,t} = 0, 1\\
    {||{\alpha}_{p,t}-\hat{\alpha}_{p,t}||_1}, &
    \text{otherwise}
    \end{cases}
\end{equation}
We treat the transition region separately from the foreground and background. 

\vspace{4pt}
\noindent{\textbf{Composition Loss.}} For the composition loss $L_c$, we calculate L1 loss of the transition region. $\hat{F}$, $\hat{B}$ and $\hat{I}$ denote groundtruth foreground, background and composited frame.
\begin{equation}\label{composition_loss}
L_c = ||\alpha_{p,t} \hat{F}_{p,t} + (1-\alpha_{p,t}) \hat{B}_{p,t} - \hat{I}_{p,t}||_{1}
\end{equation}

\vspace{4pt}
\noindent{\textbf{Gradient Loss.}} 
Let $G$ be the Sobel filter, the gradient loss $L_g$ at timestamp $t$ is defined as
\begin{equation} \label{grad_diff}
L_g = || G({\alpha}_{p,t})-G(\hat{\alpha}_{p,t})||_1 \cdot L_a
\end{equation}
Different from treating the absolute difference of gradient as a loss function, we use it as a spatial loss weight, which shares the same idea with online hard example mining.

\vspace{4pt}
\noindent{\textbf{KL-Divergence Loss.}} 
We use KL-divergence loss to provide another constraint for the alpha matte prediction against its ground truth. We first normalize ${\alpha}_{p,t}$ and $\hat{{\alpha}}_{p,t}$ by their summation value respectively, and then apply KL-divergence loss, which is defined as
\begin{equation} \label{kl_loss}
L_{kl} = D_{KL}({{\alpha}_{p,t} \over{\sum {{\alpha}_{p,t}}}},  {\hat{\alpha}_{p,t}\over{\sum\hat{\alpha}_{p,t}}})
\end{equation}
where $D_{KL}$ represents the KL-divergence function.

\vspace{4pt}
\noindent{\textbf{Temporal Coherence Loss.}} 
We enforce consistency of the predicted alpha values between consecutive frames by defining the temporal coherence loss at timestamp $t$ as
\begin{equation} \label{temp_loss}
    L_{tc} = ||\frac{\mathrm{d}{\alpha}_{p,t}}{\mathrm{d}t}-\frac{\mathrm{d}\hat{\alpha}_{p,t}}{\mathrm{d}t}||_2
\end{equation}

\noindent{\textbf{Total loss.}} Finally, the total loss is the summation of all pixels at all target timestamps, defined as
\begin{equation} \label{total_loss}
L_{total} = {1\over{\#}}\sum_{p,t}{(L_a+L_c + L_g + L_{kl} + L_{tc})}
\end{equation}
where $\#$ denotes the number of pixels.

\section{Experiments}

\subsection{Implementation Details}
\noindent{\textbf{Trimap Propagation.}} Our trimap propagation network applies two encoders adapted from ResNet-34~\cite{he2016deep} and an decoder composed of by several up-sampling and convolution layers. In the training stage, two frames are randomly sampled from a video. One is treated as the reference and the other as target. Augmentations including random cropping, coloring jittering and flipping are performed on two frames. We totally train the model for 75 epochs with a batch size of 4. The initial learning rate is set to 0.001 and then decays linearly to an end learning rate of 0.0001. Adam optimizer is applied for training all the parameters.

\noindent{\textbf{Encoder-Decoder Network.}} 
In the training stage, for each sample, we randomly select a chunk of continuous frames from the whole video as the input frames. We treat the middle frame as the target frame, and the others as neighboring frames. Then we randomly pick a $320\times 320$ patch centered on pixels in the unknown regions of the target frame and crop the $320\times 320\times (2n+1)$ cube from the chunk where $n$ is the number of neighboring frames. To make the model robust to scale variance, we crop cubes with different sizes including $320\times320$, $480\times480$, $640\times640$, and resize them to $320\times320$. Then, we apply random horizontal flipping on the cube. The above operations are conducted consistently on the composited frames and their alpha, foreground and background frames. In addition, the trimaps for the cube are randomly generated from the groundtruth alpha mattes. We dilate and erode the ground truth alpha matte with a random kernel size within a range of $[2,5]$ and a random iteration within range of $[5, 15]$.

We initialize our encoder network with the pre-trained weights on the ImageNet~\cite{imagenet} dataset and the fourth input channel with zeros. The decoder network is initialized with Xavier random variables. All models are trained for 100 epochs with batch size of 1 and $n$ set to 2. The initial learning rate is 0.00005 which is fixed in the first 20 epochs and decays linearly in the last 80 epochs with a decay rate of 0.98. We use Adam optimizer to update parameters for the whole network.

\begin{figure*}[t]
\centering 
\includegraphics[width=1.0\linewidth]{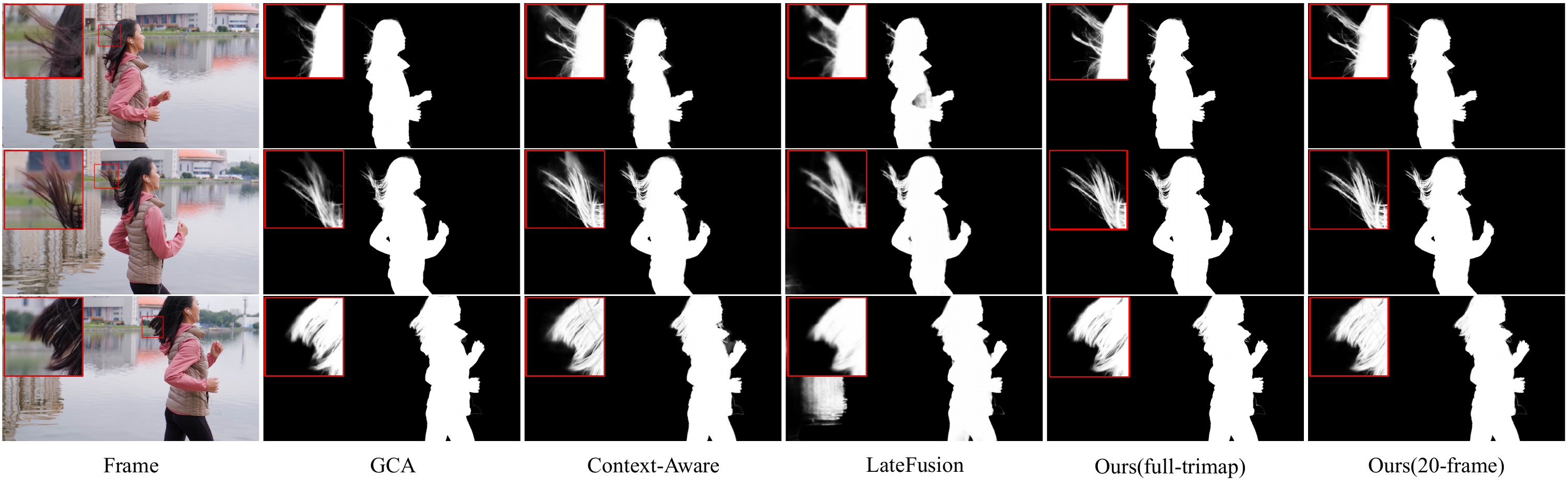}
\caption{Comparison of alpha predictions with image-based methods on real-world high-resolution videos. GCA~\cite{li2020natural} and Context-Aware~\cite{hou2019context} take frame-by-frame user-supplied trimaps as inputs. Our 20-frame model takes the trimaps propagated from the nearest reference frames provided every 20 frames.}
\label{fig:realres}
\vspace{-0.15in}
\end{figure*}

\subsection{Evaluation Metrics}
We use both image-based and video-based evaluation metrics. To evaluate the per-pixel accuracy, we follow Xu~\etal~\cite{Xu2017DeepIM} and adopt four quantitative metrics, namely the sum of absolute differences (SAD), mean square error (MSE), the gradient error (Grad) and the connectivity error (Conn). In addition, to evaluate the temporal coherency, we also take dtSSD and MESSDdt into consideration. These two metrics were proposed in \cite{erofeev2015perceptually} and defined as

\begin{equation}\label{dtssd}
\mbox{dtSSD} = {1\over{\#}}\sum_t{\sqrt{\sum_p{({d{\alpha}_{p,t}\over{dt}} - {d\hat{\alpha}_{p,t}\over{dt}})^2}}}
\end{equation}

\small
\begin{equation}
\begin{aligned}
\begin{split}\label{messddt}
\mbox{MESSDdt} =& {1\over{\#}} \sum_{p,t}|{({\alpha}_{p,t} - \hat{\alpha}_{p,t})}^2 - \\ 
&{({\alpha}_{p+v_p,t+1} - \hat{\alpha}_{p+v_p,t+1})}^2|
\end{split}
\end{aligned}
\end{equation}

Here $v_p$ denotes the motion vector at pixel $p$, which is computed by optical flow algorithm for groundtruth sequences. The evaluation code of the two temporal metrics are our own implementations.

\begin{figure}[t]
\hspace{-0.1in}\includegraphics[width=1.07\linewidth]{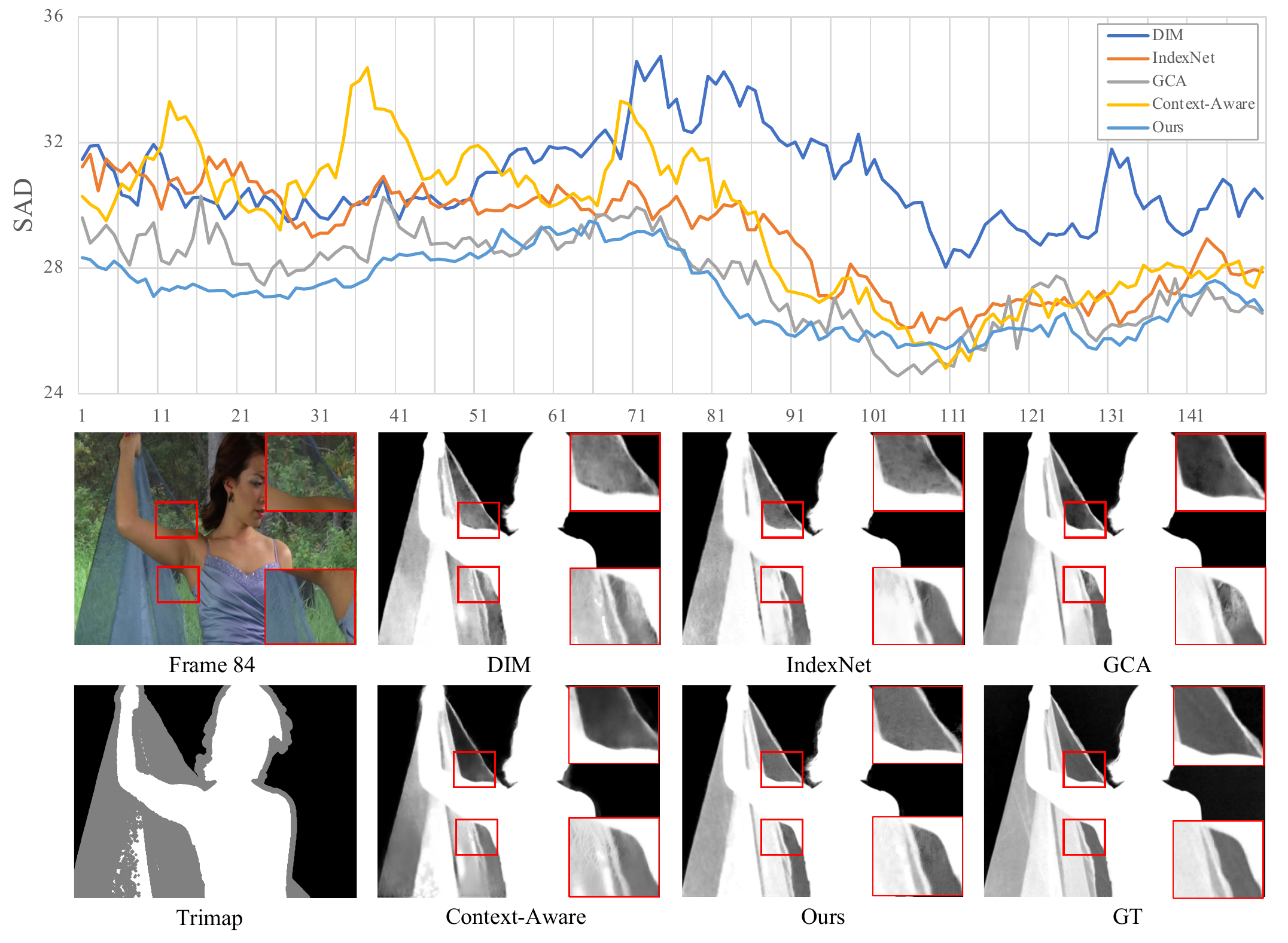} 
\caption{Comparisons with image-based methods on metric SAD.}
\label{fig:temporal}
\vspace{-5pt}
\end{figure}

\vspace{4pt}
\subsection{Results on Composited Dataset}
We evaluate our method and image-based methods on the proposed composited test set under different trimap settings, including ``full-trimap" and ``20-frame", in which user-labeled trimaps are respectively provided frame-by-frame and every 20 frames.  
Table~\ref{tab:composition} tabulates evaluation results, where our model outperforms image-based methods on all of the metrics by a large margin under dense-trimap setting. Even under the 20-frame setting, our method still achieves state-of-the-art performance. 
In addition, we also plot the quantitative results of a sample on metric
SAD frame-by-frame in Figure~\ref{fig:temporal}. Compared to image-based methods, our model generates more accurate and consistent alpha matte.

\begin{table*}[h]
    \centering
    \setlength{\tabcolsep}{4mm}{
    \begin{tabular}{l|c|ccccccc}
        \hline\hline
        Methods & Trimap Setting & SAD & MSE & Grad & Conn & dtSSD & MESSDdt\\
        \hline
        DIM~\cite{Xu2017DeepIM} & full-trimap & 54.55 & 0.030 & 35.38 & 55.16 & 23.48 & 0.53  \\
        IndexNet~\cite{lu2019indices}& full-trimap & 53.68 & 0.028 & 27.52 & 54.44 & 19.50 & 0.49 \\
        Context-Aware~\cite{hou2019context} & full-trimap & 51.78 & 0.027 & 28.57 & 49.46 & 19.37 & 0.50 \\
        GCA~\cite{li2020natural} & full-trimap & 47.49 & 0.022 & 26.37 & 45.23 & 18.36 & 0.33 \\
        Ours & full-trimap & \textbf{40.91} & \textbf{0.014} & \textbf{19.02} & \textbf{40.58} & \textbf{15.11} & \textbf{0.25} \\
        \hline\hline
        LateFusion~\cite{zhang2019latefusioncnn} & no-trimap & 69.62 & 0.042 & 45.34 & 70.70 & 38.59 & 0.71 \\
        Ours & 20-frame & 43.66 & 0.016 & 26.39 & 42.23 & 16.34 & 0.28 \\
        \hline\hline       
    \end{tabular}}
    \vspace{5pt}
    \caption{Results of our deep video matting versus image-based matting methods on the composited test set. ``full-trimap" means frame-by-frame trimaps are provided; ``no-trimap" does not use trimaps. Under ``$N$-frame" setting, trimaps are provided for every $N$th frame.}
    \label{tab:composition}
\end{table*}

\subsection{Results on Real-world High-Resolution Videos}
While experiments on our composited test set have demonstrated the our model's effectiveness, a robust matting method should generalize well to real-world videos. 
Figure~\ref{fig:realres} compares the alpha matte of our method and image-based methods on real-world videos, showing that  better foreground matte can be extracted with proper consideration of the temporal information than the image-based methods directly applied to videos. See supplementary materials for more  comparisons and  composition results.

\section{Ablation Studies}

\subsection{Effects of TFA and TFF}
ST-FAM is composed of TFA and TFF.  TFA employs deformable convolution to align features of neighboring frames. TFF applies channel-wise and spatial attention to aggregate temporal features. In order to justify the effectiveness of this design, we train a basic model without ST-FAM and then add our TFA as well as TFF in turn. Table~\ref{tab:tfa_tff} shows the comparison results. Compared to the basic model in the first row, our TFA outperforms by $1.94$. With our novel TFF module, the performance is further promoted by $1.41$. TFA and TFF modules benefit our model in aligning temporal features and discarding harmful neighboring information, which is conducive to the effective extraction of foreground objects with complex backgrounds.

\begin{table}[t]
    \centering
    \setlength{\tabcolsep}{2.5mm}{
    \begin{tabular}{p{3.2cm}|ccc}
        \hline\hline
        Method & SAD & MSE & dtSSD\\
        \hline
        basic & 44.26 & 0.016 & 16.57 \\
        basic + TFA & 42.32 & 0.015 & 15.69 \\
        basic + TFA + TFF & 40.91 & 0.014 & 15.11\\ 
        \hline\hline
    \end{tabular}}
    \vspace{5pt}
    \caption{Results of TFA and TFF.}
    \label{tab:tfa_tff}
\end{table}

\begin{table}[t]
    \centering
    \setlength{\tabcolsep}{3.5mm}{
    \begin{tabular}{c|cccccccc}
        \hline\hline
        $n$ & 1 & 2 & 3 & 4 \\
        \hline
        SAD & 48.72 & 46.98 & 46.29 & 46.30 \\
        dtSSD & 19.24 & 18.59 & 18.17 & 18.10 \\
        \hline\hline
    \end{tabular}}
    \vspace{5pt}
    \caption{Results of temporal aggregation.}
    \label{tab:nframes}
\end{table}

\begin{table}[t]
    \centering
    \setlength{\tabcolsep}{2.5mm}{
    \begin{tabular}{p{3.2cm}|ccc}
        \hline\hline
        Method & SAD & MSE & dtSSD\\
        \hline
        naive-fusion & 44.13 & 0.016 & 16.62 \\
        cross-attention-fusion & 41.29 & 0.015 & 15.37 \\
        flow-fusion & 44.06 & 0.016 & 16.34 \\
        ST-FAM (Ours) & \textbf{40.91} & \textbf{0.014} & \textbf{15.11} \\
        \hline\hline
    \end{tabular}}
    \vspace{5pt}
    \caption{Results of different temporal fusion networks.}
    \label{tab:temporal_fusion}
\end{table}

\begin{table}[t]
    \centering
    \setlength{\tabcolsep}{2.5mm}{
    \begin{tabular}{p{3.2cm}|ccc}
        \hline\hline
        Trimap Setting & SAD & MSE & dtSSD\\
        \hline
        full-trimap & 40.91 & 0.014 & 15.11 \\
        20-frame & 43.66 & 0.016 & 16.34 \\
        40-frame & 52.85 & 0.026 & 19.23 \\
        1-trimap & 65.33 & 0.039 & 35.46 \\ 
        \hline\hline     
    \end{tabular}}
    \vspace{5pt}
    \caption{Results of different trimap settings.}
    \label{tab:trimap_setting}
    \vspace{-3pt}
\end{table}

\subsection{Effects of Temporal Aggregation}
Temporal information from neighboring frames help the model distinguish foreground and background pixels. The context information along temporal dimension also has an impact on the aggregation, which is decided by the number of neighboring frames in our framework. To exploit ST-FAM's ability of aggregating temporal information, we conduct experiments on models with different $n$ neighboring frames. 
Table~\ref{tab:nframes} shows that when we increase $n$ before the saturation point, the model learns temporal information from more adjacent frames so that it has a better understanding of object's motion and give more accurate and consistent predictions of alpha mattes.

\subsection{Effects of Temporal Fusion Network}
Making use multiple frames constitutes the main difference between video-based methods and image-based methods. To exploit an effective design of aggregating temporal information in matting task, we compare several temporal fusion networks, including naive-fusion, cross-attention-fusion, and flow-fusion. 
Naive-fusion aggregates temporal information through several $3\times3$ convolution layers, while cross-attention-fusion applies a cross-attention based correlation layer to compute the similarities between two frames. Flow-fusion obtains the motion vector of pixels via a lightweight flow estimation network, and the estimated motion vector is concatenated with features from the decoder for final predictions. More implementation details are provided in supplementary materials.

Table~\ref{tab:temporal_fusion} shows the quantitative comparisons. Although the cross-attention-fusion is also capable of aggregating temporal information, the cost of computation increases rapidly when more neighboring frames are integrated.

\subsection{Effects of Trimap Propagation}
To evaluate the performance of our trimap propagation method, we compare our matting results under different trimap settings: full-trimap mode, $N$-frame mode  ($N>1$), and $1$-trimap mode. In the full-trimap mode frame-by-frame trimaps are provided by users; in the $N$-frame mode user-supplied trimaps are provided for every $N$th frame; in $1$-trimap mode only one user-supplied trimap at $t=0$ is provided for the entire video. For each target frame, its nearest neighbor is found in temporal domain that contains the user-supplied trimap as the reference frame.

Table~\ref{tab:trimap_setting} tabulates the quantitative results of the different trimap settings. We can see that even under the 20-frame setting our performance only drops slightly. 
Qualitative results can be found in supplementary materials.

\section{Conclusion}
This paper proposes a new deep video matting framework that exploits temporal information between the target and reference as well as neighboring frames. This framework consists of an encoder-decoder structure using novel spatio-temporal feature aggregation modules. The proposed module benefits our model in enhancing temporal coherence leading to significantly better alpha prediction in objects with rapid motions or complex backgrounds.
This paper also contributes a large-scale video matting dataset that covers a great variety of unique matting cases to complete the data gap in present and future deep video matting research. 
We have conducted extensive experiments on our proposed test set and real-world high-resolution videos to validate our method on dealing with complex scenes.

{\small
\bibliographystyle{ieee_fullname}
\bibliography{egbib}
}

\end{document}